%% file: main.tex
\documentclass[runningheads, dvipsnames,table,xcdraw]{llncs}

\usepackage{wrapfig}
\usepackage{eccv}

\usepackage{eccvabbrv}
\usepackage{graphicx}
\usepackage{booktabs}
\usepackage{tabularx}
\usepackage[accsupp]{axessibility}  %

\usepackage[breaklinks,colorlinks,citecolor=eccvblue]{hyperref}
\usepackage{hyperref}

\usepackage{orcidlink}

\input{header}
\begin{document}
\title{Bridging the Gap: Studio-like Avatar Creation from a Monocular Phone Capture}

\author{ShahRukh Athar\(^{\text{1,3}}\)\thanks{Work done while interning at Meta Reality Labs} 
\hspace{0.3cm} Shunsuke Saito\(^{\text{2}}\) \hspace{0.3cm}
Zhengyu Yang\(^{\text{2}}\) \hspace{0.3cm}
Stanislav Pidhorskyi\(^{\text{2}}\) \hspace{0.3cm} Chen Cao\(^{\text{2}}\) \\~\\
\(^{\text{1}}\)\text{Captions Research, New York} \hspace{0.7cm} \(^{\text{2}}\)\text{Meta Reality Labs, Pittsburgh.} \\~\\ \hspace{0.7cm} \(^{\text{3}}\)\text{Stony Brook University, New York}
\institute{\email{shahrukh@nocapinc.com} \hspace{0.3cm} \email{shunsuke.saito16@gmail.com} \hspace{0.3cm}  \email{stpidhorskyi@meta.com} \hspace{0.3cm} \email{zhengyu-yang@outlook.com} \hspace{0.3cm}  \email{zju.caochen@gmail.com}}
}

\authorrunning{Athar et al.}
\maketitle
\begin{figure*}[h!]
\centering
\includegraphics[width=0.99\linewidth]{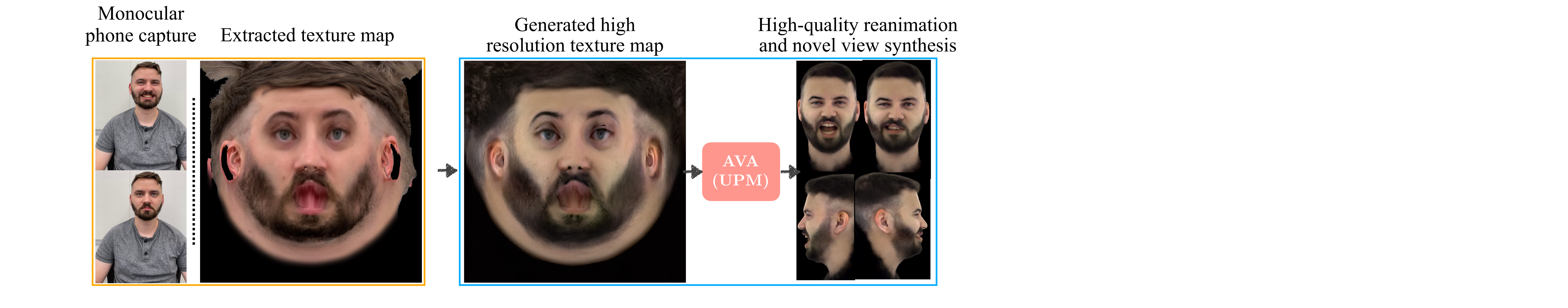}
\captionof{figure}{\small{
From a low-resolution texture map obtained through monocular phone capture, our method produces a high-resolution texture map with precise facial details, studio-like illumination, and inpainted missing regions. This generated texture map can subsequently be utilized to create a high-quality, photorealistic avatar using the pretrained Universal Prior Model (UPM) from AVA \cite{cao2022authentic}.}}

\label{fig:teaser}
\end{figure*}
\vspace{-0.75cm}
\input{00_abstract}
\input{01_intro}
\input{02_related}

\input{03_method}

\input{04_results}

\input{10_conclusion}

{\small
\bibliographystyle{splncs04}
\bibliography{11_references}
}

\newpage
\end{document}

%% file: header.tex
\input{macros}

\usepackage[capitalize]{cleveref}
\crefname{section}{Sec.}{Secs.}
\crefname{table}{Table}{Tables}
\crefname{figure}{Fig.}{Figs.}

%% file: macros.tex
\usepackage{graphicx}	
\usepackage{amsmath}	
\usepackage{amssymb}	
\usepackage{booktabs}
\usepackage{times}
\usepackage{microtype}
\usepackage{epsfig}
\usepackage{xcolor}
\usepackage{caption}
\usepackage{float}
\usepackage{placeins}
\usepackage{color, colortbl}
\usepackage{stfloats}
\usepackage{enumitem}
\usepackage{tabularx}
\usepackage{xstring}
\usepackage{multirow}
\usepackage{xspace}
\usepackage{url}
\usepackage{subcaption}
\usepackage[hang,flushmargin]{footmisc}
\usepackage{comment}
\usepackage{mathalpha}
\usepackage{float}
\usepackage{cuted}
\usepackage{soul}

\usepackage{multirow}

\usepackage{tabularx}
\usepackage{mathtools}
\usepackage{booktabs}
\usepackage{enumitem}%
\usepackage{verbatim}
\usepackage{subcaption}
\usepackage{amsfonts} %

\newcommand{\R}[1]{{%
    \textbf{%
        \ifstrequal{#1}{1}{\textcolor{red}{R#1}}{%
        \ifstrequal{#1}{2}{\textcolor{blue}{R#1}}{%
        \ifstrequal{#1}{3}{\textcolor{magenta}{R#1}}{%
        \ifstrequal{#1}{4}{\textcolor{teal}{R#1}}{%
                           \textcolor{cyan}{R#1}%
        }}}}%
    }%
}}

\DeclareMathSymbol{\shortminus}{\mathbin}{AMSa}{"39}

\newcommand{\fig}[1]{Fig~\ref{fig:#1}}
\newcommand{\sect}[1]{Sect~\ref{sect:#1}}

\newcommand{\tab}[1]{Table~\ref{tab:#1}}

\newcommand{\eq}[1]{Eq. (\ref{eq:#1})}

\newcommand{\Gitw}{\mathcal{G}_{phone}}
\newcommand{\GMug}{\mathcal{G}_{Studio}}
\newcommand{\Iitw}{\mathbf{\mathcal{I}}^{phone}}
\newcommand{\IMug}{\mathbf{\mathcal{I}^{*}}}
\newcommand{\IGTM}{\mathbf{{\mathcal{I}}}^{Studio}}
\newcommand{\IGTMgr}{\mathbf{{\mathcal{I}}}^{Studio*}}

\definecolor{turquoise}{cmyk}{0.65,0,0.1,0.3}
\definecolor{purple}{rgb}{0.65,0,0.65}
\definecolor{dark_green}{rgb}{0, 0.5, 0}
\definecolor{orange}{rgb}{0.8, 0.6, 0.2}
\definecolor{red}{rgb}{0.8, 0.2, 0.2}
\definecolor{darkred}{rgb}{0.6, 0.1, 0.05}
\definecolor{blueish}{rgb}{0.0, 0.3, .6}
\definecolor{light_gray}{rgb}{0.7, 0.7, .7}
\definecolor{greyblue}{rgb}{0.25, 0.25, 1}
\definecolor{drivevid}{RGB}{75,154,250}
\definecolor{meshrig}{RGB}{241,135,40}
\definecolor{nvs}{RGB}{236,71,250}

\usepackage{blindtext}

\renewcommand{\paragraph}[1]{\vspace{1em}\noindent\textbf{#1}.}
\newcommand{\mathcolorbox}[2]{\colorbox{#1}{$\displaystyle #2$}}
\usepackage{environ}
\NewEnviron{smequation}{%
    \begin{equation}
    \scalebox{0.90}{$\BODY$}
    \end{equation}
}

%% file: 00_abstract.tex
\begin{abstract}
\vspace{-0.3cm}
Creating photorealistic avatars for individuals traditionally involves extensive capture sessions with complex and expensive studio devices like the LightStage system. While recent strides in neural representations have enabled the generation of photorealistic and animatable 3D avatars from quick phone scans, they have the capture-time lighting baked-in, lack facial details and have missing regions in areas such as the back of the ears. Thus, they lag in quality compared to studio-captured avatars.
In this paper, we propose a method that bridges this gap by generating studio-like illuminated texture maps from short, monocular phone captures. We do this by parameterizing the phone texture maps using the $W^+$ space of a StyleGAN2, enabling near-perfect reconstruction. Then, we finetune a StyleGAN2 by sampling in the $W^+$ parameterized space using a very small set of studio-captured textures as an adversarial training signal. To further enhance the realism and accuracy of facial details, we super-resolve the output of the StyleGAN2 using carefully designed diffusion model that is guided by image gradients of the phone-captured texture map.
Once trained, our method excels at producing studio-like facial texture maps from casual monocular smartphone videos. Demonstrating its capabilities, we showcase the generation of photorealistic, uniformly lit, complete avatars from monocular phone captures. \href{http://shahrukhathar.github.io/2024/07/22/Bridging.html}{The project page can be found here.}
\end{abstract}

%% file: 01_intro.tex
\vspace{-0.2cm}
\section{Introduction}
\label{sec:intro}
Photorealistic and animatable avatars are paramount for lifelike human-to-human interactions in AR/VR applications. Creating high-fidelity avatars often requires sophisticated devices like the LightStage capture system that capture human heads with a range of facial expressions \cite{lombardi2018deep, lombardi2021mixture}. Typically, these captures occur in a studio-like environment, with uniform illumination and densely sampled views to reconstruct complete avatars with consistent lighting.
While these methods excel in generating hyper-realistic avatars, they cannot be scaled to millions of people as capturing so many people with a LightStage-like capture system is impractical and non-trivial.  

Recently, there has been extensive work in generating photorealistic avatars through a monocular capture \cite{cao2022authentic, grassal2022neural, rignerf}. These methods involve scanning various facial expressions and head poses of the user to reconstruct a 3D avatar that closely aligns with the captured data. However, a notable limitation of these methods is that the captured lighting is embedded in the avatars, causing their appearance to be heavily dependent on the capture devices and surrounding environments.
Furthermore, due to the constraints of the capture setup, certain areas, such as the back of the head or ears, are never visible, leaving visible holes and artifacts when viewed during animation.

An innovative strategy to tackle this challenge is to use image-to-image translation to transform the phone data into studio-captured data. This transformation can be learnt through supervised training using paired data \cite{pix2pix2016}, or through unsupervised training using unpaired datasets \cite{zhu2017cycleGAN}. Due to reasons mentioned earlier, creating a large-scale paired dataset of studio and phone captures is impractical, which rules out the possibility of using supervised image-to-image translation methods. On the flip side, current unsupervised image-to-image translation methods fail to preserve fine detail in the transformation process \cite{song2021agilegan, zhu2017cycleGAN} which is paramount to creating a high-fidelity avatar.

In this paper, we introduce a method capable of generating studio-like, high-quality avatars from monocular phone captures. We do this by parameterizing a large-scale dataset of phone-captured face texture maps using the \(W^{+}\) space of a StyleGAN2. Then, we finetune this StyleGAN2 by sampling from this parametrized \(W^{+}\) space using a small set of unpaired studio-captured texture maps, to create a Studio-StyleGAN2 model. Our key insight is that sampling from the \(W^{+}\) space instead of Z space leads to a more generalizable model as samples from the \(W^{+}\) are, by construction, as diverse as the training set while samples from the Z space often suffer from mode-collapse. %
During inference, the given phone-captured texture map, is first inverted to the \(W^{+}\) space of the StyleGAN2. This inverted \(W^{+}\) vector is then given as input to the Studio-StyleGAN2 to generate a low-resolution studio-like texture map. Finally, a novel facial detail conditioned diffusion model is used to enhance facial details of the low-resolution studio-like texture map obtained in the previous step. Once trained, our method excels at producing high-quality studio-like face texture maps from monocular phone captures, which are then used as inputs to the universal face prior model (UPM) from Authentic Volumetric Avatars (AVA) \cite{cao2022authentic} to generate photorealistic avatars.
\noindent In summary, our key contributions are as follows:
\begin{itemize}
\item Introducing a groundbreaking method for creating studio-like, photorealistic avatars from monocular phone captures. 
\item Innovative finetuning of a pre-trained generative model using a minimal dataset from another domain by sampling in the inverted $W^+$ space, which facilitates the development of a generative model for a new domain while preserving the integrity of the $W^+$ latent space.
\item A novel diffusion model conditioned on the phone texture gradient, that is designed to upsample studio textures, effectively enhancing facial details and contributing to the overall realism of the generated avatars.
\end{itemize}

%% file: 02_related.tex
\section{Related Work}
\label{sec:related}

Our paper aims to bridge the gap between a studio capture and a phone capture by creating a studio-like, photorealistic avatar through a monocular phone capture. In delineating the related work, we offer a comprehensive overview of key research domains, including studio-captured avatars, phone-captured avatars, and image-to-image translation.

\textit{Studio-captured avatars.} 
The reconstruction of high-fidelity static and dynamic models of the human head based on photometric measurements has a long-standing history in computer graphics and vision. Achieving a photorealistic human avatar often requires specialized hardware in high-end production, such as the LightStage system. To model the complex skin appearance, various physically-based models have been explored. Notably, subsurface scattering \cite{borshukov2005realistic}, linear polarization patterns \cite{ghosh2011multiview} and  fine-scale skin details \cite{alexander2009digital, alexander2013digital} have been investigated.
For dynamic expression details, Jimenez et al. \cite{jimenez2010practical} compute dynamic skin appearances by blending hemoglobin distributions captured with different expressions. In their subsequent work \cite{jimenez2018practical}, expression-dependent normal maps are interpolated to add realistic wrinkles to an animated face. 
Nagano et al.\cite{nagano2015skin} synthesize skin microstructures based on local geometric features derived from high-precision microgeometry, acquired with an LED sphere and a skin deformer.
While these methods have been instrumental for offline movie production, their substantial compute requirements make them less suitable for real-time applications. Despite recent efforts to enable real-time rendering of physically-based avatars \cite{seymour2017meet}, heavy compute remains challenge.
In response to the challenges posed by complex physical computations, researchers have proposed a deep appearance model \cite{lombardi2018deep}. This model utilizes a coarse 3D triangle mesh in conjunction with view-dependent texture mapping. The texture is regressed by a neural network conditioned on viewpoint and expression latent codes. This conditioning accounts for view- and expression-dependent variations while compensating for the imperfect proxy geometry. Subsequent work extends the mesh-texture representation to a volumetric representation using a Mixture of Volumetric Primitives (MVP) \cite{lombardi2021mixture}, further enhancing the model's quality.
Pixel Codec Avatars (PiCA), as demonstrated by \cite{ma2021pixel}, showcase the efficiency of rendering such models, even on mobile hardware platforms, by leveraging efficient per-pixel processing. Moreover, Bi et al. \cite{bi2021deep}, based on relighting captured data, can also relight avatars with any novel point light or environment maps.
While these methods can achieve hyper-realistic avatars, their studio requirements make them challenging to generalize to ordinary users.

\textit{Phone-captured avatars.}
There are several methods aimed at creating avatars in lightweight ways, even from a phone or a single image. Some of these approaches focus on generating stylized avatars based on different input requirements, ranging from capturing multi-view images using a phone \cite{ichim2015dynamic} to utilizing a single monocular image \cite{shi2019face,luo2021Normalized,lin2021meingame,sang2022agileavatar}. While these methods excel in producing animatable avatars, it is important to note that their appearance tends to be cartoonish and lacks realism.

Another category of methods is dedicated to creating realistic human avatars based on graphics pipelines \cite{casas2015rapid, cao2016real, garrido2016reconstruction, hu2017avatar, yamaguchi2018high, lattas2020avatarme, lattas2021avatarme++}. While relying on traditional graphics methods, these approaches often result in avatars that appear uncanny.
Building upon deep learning-based representations, researchers have proposed methods to generate avatars with increased realism, including notable works by Gafni et al. \cite{gafni2021dynamic} and Grassal et al. \cite{grassal2022neural}. Among these, the work of Cao et al. \cite{cao2022authentic}, Authentic Volumetric Avatars (AVA), stands out for its focus on creating photorealistic avatars from phone scans. The process involves training a Universal Prior Model (UPM) using studio-captured data, followed by personalizing this UPM using data from a phone scan of an unseen subject. While the method successfully produces avatars with realistic appearance and animation, it is worth noting that the lighting is baked into the personalized avatar, and certain details, such as the back of the ears, may be missing.

\textit{Image-to-image translation}
We use image-to-image translation to map images from the source domain (phone data) to the target domain (studio data). Isola et al. introduced Pix2Pix \cite{pix2pix2016}, a method that utilizes adversarial training strategies \cite{goodfellow2020generative} to achieve this mapping. Additionally, Wang et al. \cite{wang2018high} focused on increasing the resolution of generated results from semantic label maps. While these methods successfully map images between domains, it is essential to note that they require paired training data for effective implementation.
In many real-world scenarios, obtaining paired training data can be challenging and expensive. 

To address this issue, unsupervised image-to-image translation has been introduced. Zhu et al. \cite{zhu2017cycleGAN} proposed a novel cycle-consistency loss to ensure that translating an image from one domain to another and back again should result in the original image. This helps the model maintain consistency and produce more realistic translations. Subsequent methods have further improved unsupervised image-to-image translation from different perspectives, including a multimodal model \cite{huang2018multimodal}, few-shot input for video-to-video translation \cite{wang2019few}, translation of images with human control \cite{pinkney2020resolution}, and translation of real images into different styles \cite{song2021agilegan}. However, none of these methods are designed with preservation of facial identity and the generation of realistic facial details in mind, especially with such little training data. Later in the paper we show how this prior work compares to ours for transforming low-resolution phone-captured texture maps to high-resolution studio-captured texture maps.

%% file: 03_method.tex
\section{Method}
\label{sec:method}
\begin{figure*}[h!]
    \centering
    \includegraphics[width=1.01\linewidth]{figs/Method_mod.pdf}
    
    \caption{\small{\textbf{Method Overview.} 
    Our method employs a two-step process to train \(\GMug\). Initially, we train a StyleGAN2 on 12k neutral texture maps captured by phones, yielding \(\Gitw\). Subsequently, we initialize \(\GMug\) with the weights of \(\Gitw\) and fine-tune it by sampling from \(S_{W^{+}} = \{W^{+}_{\Iitw_{0}}, \hdots, W^{+}_{\Iitw_{N-1}}\}; \), where \(N = 12k\), and \(W^{+}_{\Iitw_{i}}\) represents the vector obtained by inverting the \(i\)'th phone-captured texture map, \(\Iitw_{0}\), in the \(W^{+}\) space of \(\Gitw\).
    During inference, the given phone-captured texture map is inverted to \(W^{+}\), then passed to \(\GMug\) to generate a low-resolution studio-lit texture map, \(\IMug\). Accurate facial details are subsequently added using diffusion model \(f_{\phi}\), conditioned on the gradient of the phone texture. This process results in the final high-resolution, studio-lit, and completed texture map, \(\IMug\).   
    }}
    \label{fig:method}
\end{figure*}

In this section, we present our method for generating studio-like avatars from the phone captures. Our approach consists of two key components: a StyleGAN2 for texture translation and a diffusion model for facial detail generation.
In \sect{method_stylegan}, we describe the generation of a low-resolution texture map with studio-like lighting and missing regions inpainted. First, we pretrain a StyleGAN2 on 12k phone textures and then finetune it using a small set of studio-captured texture maps. In order to improve the generalization of the finetuned StyleGAN2, we optimize it by  sampling in the \(W^{+}\) space, instead of the W-space or Z-space, using a set of 12k \(W^{+}\) vectors obtained by inverting the phone captured texture maps.   
In \sect{method_diffusion_model}, we introduce a diffusion model that generates facial details. The diffusion model takes the output from the aforementioned StyleGAN2 and generates a high-resolution neutral texture with realistic facial details.
After obtaining a high-resolution studio-like neutral texture from our method, we use it to learn a color transform to transfer phone-captured expression textures to studio-lit expression textures. These expressions are subsequently utilized for animating a high-quality avatar (\sect{method_driving}). Notably, we intentionally avoid applying inpainting or super-resolution techniques to the expression textures. 
Fig. \ref{fig:method} provides an overview of our method. 

\subsection{Illumination manipulation and inpainting}
\label{sect:method_stylegan}

In our initial step, we track the geometry from a monocular phone capture of the user's neutral face and extract the neutral texture \(\Iitw\) from the captured image, employing the method outlined in \cite{cao2022authentic}. Subsequently, we translate this phone-captured texture with in-the-wild lighting and missing regions into a texture map with studio lighting and missing regions in-painted. This translation is accomplished by parametrizing \(\Iitw\) using the \(W^{+}\) space of a StyleGAN2, as follows:

\begin{smequation}
\begin{split}
    W^{+}_{\Iitw} & = \underset{W^{+}}{\text{argmin}} ~||\Gitw(W^{+}) - \Iitw||_{2}^{2}  + LPIPS(\Gitw(W^{+}), \Iitw),
\end{split}    
\label{eq:wp_inv}
\end{smequation}
where \(\Gitw\) is a StyleGAN2 trained on phone-captured textures, and \(W^{+}_{\Iitw}\) is the optimized vector in the \(W^{+}\) space of \(\Gitw\). Now, we generate the low-resolution studio-like texture as follows:
\begin{smequation}
    \IMug = \GMug(W^{+}_{\Iitw}),
    \label{eq:mug_f}
\end{smequation}
where \(\GMug\) is a StyleGAN2 responsible for generating low-resolution studio-like textures, while \(\IMug\) refers to the studio-lit version of \(\Iitw\) with the inpainted missing regions. Ideally, \(\IMug\)  should retain the identity and semantics of \(\Iitw\), with the only distinctions being the illumination and the inpainting of missing regions. Below we describe the training procedure for \(\GMug\).\\
\indent We initialize the synthesis network of \(\GMug\) using weights from \(\Gitw\), which we then finetune to generate texture maps with studio lighting and inpaint the missing regions. We choose to finetune \(\Gitw\) for the following reasons: 
1) Due to the immense costs of a studio capture, studio quality training data is typically very limited. In our case, we only have 383 studio-captured texture maps, making training from scratch hard \cite{StyleGANADA}; 2) Since we want \(\IMug\) and \(\Iitw\) to have the same facial identity and semantics meaning, it is beneficial to learn a latent space that is shared between \(\GMug\) and \(\Gitw\). For finetuning, we first invert the entire dataset of \(N\) phone-captured texture maps using \eq{wp_inv}, giving us a set of \(N\) vectors in the \(W^{+}\) space: \(S_{W^{+}} = \{W^{+}_{\Iitw_{0}}, \hdots, W^{+}_{\Iitw_{N-1}}\}\).
Next, we randomly sample vectors from the inverted set, \(S_{W^{+}}\), and finetune the synthesis network of \(\Gitw\) to give \(\GMug\). 
We choose to finetune the generator with samples from inverted \(W^{+}\) instead of the traditional \(W\) space, since they are, by construction, from the real data distribution and are consequently more diverse, leading to better generalization of \(\GMug\). The losses we use during finetuning are described as below.\\
\indent \textit{Studio-Discriminator loss.} This loss uses a discriminator to ensure the texture maps generated by \(\GMug\) are from the distribution of studio-captured texture maps:
\vspace{-0.2cm}
\begin{smequation}
\begin{split}
\mathcal{L}_{Adv} &= \mathbb{E}_{\IGTM \sim P(\IGTM)}\left[min(0, -1 + \mathcal{D}_{Studio})\right] \\
&+ \mathbb{E}_{W^{+}_{\Iitw_{i}} \sim S_{W^{+}}}\left[min(0, -1 -\mathcal{D}_{Studio}(\GMug(W^{+}_{\Iitw_{i}}))\right],
\end{split}
\end{smequation}
where \(\mathcal{D}_{Studio}\) is a discriminator that initialized from the pretrained phone-captured texture StyleGAN2 Discriminator and finetuned using ground-truth studio-captured textures and textures generated by  \(\GMug\).

\textit{Face-Identity Loss} This loss compares the face identity embeddings of the renders of texture maps generated from \(\GMug\) and \(\Gitw\) in order to ensure the identity is preserved:
\vspace{-0.2cm}
\begin{smequation}
\mathcal{L}_{FaceID} = \left\||\mathcal{F}\left(\GMug(W^{+}_{\Iitw_{i}})\right) - \mathcal{F}\left(\Gitw(W^{+}_{\Iitw_{i}})\right)\right\||_{2}^{2},
\label{eq:faceid}
\end{smequation}
where \(\mathcal{F}\) is a pretrained face recognition network\footnote{We use the network from \href{https://github.com/timesler/facenet-pytorch}{here}}.

\textit{Perceptual Loss} This loss encourages the preservation of semantic features between the texture maps generated \(\GMug\) and \(\Gitw\) by minimizing their distance in the VGG feature space
\vspace{-0.2cm}
\begin{smequation}
\mathcal{L}_{Percp} = \text{LPIPS}\bigg(\GMug(W^{+}_{\Iitw_{i}}), \Gitw(W^{+}_{\Iitw_{i}})\bigg).
\end{smequation}

\textit{Perceptual Reconstruction Loss} This loss ensures that skintones are preserved using a small amount of paired data (i.e subjects for whom we have both phone and studio textures maps i.e both \(\Iitw\) and \(\IGTM\)) 
\vspace{-0.2cm}
\begin{smequation}
\mathcal{L}_{Percp-Recons} = \frac{1}{k}\sum_{i = 1}^{K}\text{LPIPS}\bigg(\GMug(W^{+}_{\Iitw_{i}}), \IGTM_{i}\bigg),
\label{eq:percept_recons}
\end{smequation}
where \(\IGTM_{i}\) and  \(\Iitw_{i}\) are the ground-truth studio and ground-truth phone texture maps of the same person. We only have a very small amount of this data for training (K=83).
\textit{R1-Regularization}: We regularize \(\mathcal{D}_{Studio}\) using the R1 regularization \cite{mescheder2018training} as follows:
\vspace{-0.2cm}
\begin{smequation}
    \mathcal{L}_{R1} = -\frac{\gamma}{2}\mathbb{E}_{\IGTM \sim P(\IGTM)}\left[||\nabla \mathcal{D}_{Studio}(\IGTM)||_{2}^{2}\right],
\end{smequation}
where \(\gamma = 10\).

In order to further encourage identity preservation, we leverage the multi-resolution architecture of StyleGAN2 \cite{StyleGAN} and only optimize the generator parameters after the \(8\times8\) resolution during the finetuning process. The intuition being that identity-specific information is mostly stored in the low-resolution maps, we ablate this choice in the Supplementary section of the paper. The final optimization for the joint finetuning of \(\GMug\) and \(\mathcal{D}_{Studio}\) is the following:
\vspace{-0.2cm}
\begin{smequation}
\underset{\GMug^{\theta(8+)}}{\text{min}} \underset{\mathcal{D}_{Studio}}{\text{max}}\mathcal{L}_{Adv} +  \mathcal{L}_{R1} + \mathcal{L}_{Percp-Recons} +\lambda_{1}\mathcal{L}_{Percp} + \lambda_{2}\mathcal{L}_{FaceID},
\end{smequation}
where \(\lambda_{1} = 0.5\) and \(\lambda_{2} = 10\). We ablate \(\mathcal{L}_{Percp-Recons}, \mathcal{L}_{Percp}, \mathcal{L}_{FaceID}\) in the supplementary.

\subsection{Synthesis of accurate facial details}
\vspace*{-3mm}
\label{sect:method_diffusion_model}
\paragraph{Generating high-quality facial details} \(\GMug\) can produce an inpainted texture map, \(\IMug\), with illumination consistent with that of a studio capture. However, due to the limited amount of finetuning data, \(\GMug\) often struggles to reproduce facial details that are visibly present in \(\Iitw\) itself. This limitation results in oversmoothed and inaccurate avatars. Achieving realistic facial details necessitates an accurate modeling of the facial texture distribution, a task that the StyleGAN2 fails to do.

Motivated by the recent success of latent diffusion models in accurately modeling data distributions, we propose the task of generating facial details as the reverse process of a Markov chain that transforms a low-resolution face texture map to a high-resolution one. We adopt the formulation from \cite{yue2023resshift}, where the residual between the low-res and high-res studio-captured texture maps, denoted as \(e_{0} = \IGTM_{LR} - \IGTM\), is used to define the Markov process. More specifically, the forward process is defined as follows:
\vspace{-0.3cm}
\begin{smequation}
    \begin{split}
    q(\IGTM_{t}|\IGTM_{t-1}, \IGTM_{LR}) = \mathcal{N}&\bigg(\IGTM_{t}; \IGTM_{t-1} + \alpha_{t}e_{0}, \kappa^{2}\alpha_{t}\mathbf{\textit{I}}\bigg),  \\
    & t=1,2,\hdots, T,
    \end{split}
    \label{eq:f_proc}
\end{smequation}
where \(\alpha_{t}\) controls the schedule with which the residual is added, \(\kappa\) is a hyperparameter controlling the noise schedule and \(\mathbf{\textit{I}}\) is the identity matrix. The reverse process is defined as 
\vspace{-0.2cm}
\begin{smequation}
    \begin{split}
    q(&\IGTM_{t-1}|\IGTM_{\{t, LR\}}, \IGTM) = \mathcal{N}\Bigg(\IGTM_{t-1}\bigg| \frac{\eta_{t-1}}{\eta_{t}}\IGTM_{t} + \frac{\alpha_{t}}{\eta_{t}}\IGTM, \kappa^{2}\frac{\eta_{t-1}}{\eta_{t}}\alpha_{t}\mathbf{\textit{I}}\Bigg),
    \end{split}
    \label{eq:re_proc}
\end{smequation}
where \(\eta_{t} = \alpha_{t} + \eta_{t-1}\). We refer the reader to \cite{yue2023resshift} for details regarding the formulation of both the forward and reverse process. Finally, the diffusion model, \(f_{\phi}\), is trained to minimize the following objective
\vspace{-0.2cm}
\begin{smequation}
    \underset{\phi}{\text{min }} \sum_{t}||f_{\phi}(\IGTM_{t}, \IGTM_{LR}, t) - \IGTM||_{2}^{2}
    \label{eq:diff_obj}
\end{smequation}
During inference, we use \(f_{\phi}\) to add realistic facial details to the low-resolution output of \(\GMug\) as follows:
\vspace{-0.2cm}
\begin{smequation}
    \hat{\IMug} = \text{ReverseProcess}\left(f_{\phi}, \IMug\right).
    \label{eq:diff_inf}
\end{smequation}
\vspace{-0.3cm}
\paragraph{Recovering accurate facial details} 
While \(f_{\phi}\) enhances \(\IMug\) by adding realistic facial details, it struggles to recover details already present in the phone-captured texture map \(\Iitw\) due to the inherent lack of such details in 
\(\IMug\). The loss of details occurs during the illumination manipulation and inpainting process, as described in \eq{wp_inv} and \eq{mug_f}. To address this issue, we incorporate the image gradient from the phone-captured texture map into the low-resolution texture map during the training of the diffusion model, as follows:
\vspace{-0.2cm}
\begin{smequation}
    \IGTMgr_{LR} = \IGTM_{LR} + \mathbb{G}(\Iitw)
\end{smequation}
where \(\Iitw\) represents the phone-captured texture map, and \(\IGTM\) is the studio-captured texture map of the same person. $\mathbb{G}$ denotes the operator used to calculate the image gradient. During training, \(\IGTMgr_{LR}\) replaces \(\IGTM_{LR}\) in \eq{f_proc}, \eq{re_proc}, and \eq{diff_obj}. In the inference stage, we augment \(\IMug\) by adding the gradient of the phone-captured texture map as follows:
\vspace{-0.2cm}
\begin{smequation}
    \hat{\IMug} = \text{ReverseProcess}\left(f_{\phi}, \IMug + \mathbb{G}(\Iitw)\right)
    \label{eq:diff_inf_v2}
\end{smequation}

\textit{Implementation details} 
We utilize 83 paired texture maps, representing subjects for whom we have both \(\IGTM\) and \(\Iitw\), to train the diffusion model. Following the approach in \cite{yue2023resshift}, we employ a latent diffusion model. To prevent overfitting, training is conducted on random \(512 \times 512\) crops of the \(1024 \times 1024\) resolution texture map. During the inference stage, we employ the full-resolution texture map.

\subsection{Driving a high-quality avatar}
\label{sect:method_driving}
With the studio-lit version \(\IMug\) now available for a given neutral phone-captured texture map \(\Iitw\), we proceed to estimate a color transform mapping from the phone-captured texture map to the studio-lit texture map as follows:
\vspace{-0.2cm}
\begin{smequation}
    \{G, B\} \leftarrow \underset{\{G, B\}}{\text{argmin }} ||\IMug - \left(\texttt{Rsz}(G)\times \Iitw + \texttt{Rsz}(B)\right)||.
    \label{eq:color_transform}
\end{smequation}
Here, \(G\) and \(B\) represent gain and bias maps of resolution \(k \times k\) and \texttt{Rsz} is the resizing operator with bilinear interpolation. Utilizing \(G\) and \(B\), we perform a transformation on phone-captured expression texture maps to achieve studio-like lighting, as outlined below:
\begin{smequation}
    \IMug_{exp} = \texttt{Rsz}(G) \times \Iitw_{exp} + \texttt{Rsz}(B).
    \label{eq:color_transform_tex}
\end{smequation}
In our experiment, we select the value of $k=32$ to efficiently transform the lighting while preserving the details.

Now, the studio-like high-resolution neutral texture $\hat{I^*}$ serves as conditioning data for the universal avatar prior from \cite{cao2022authentic}. By combining the expression code generated by inputting $\IMug_{exp}$ into the expression encoder from \cite{cao2022authentic}, we can render a high-quality avatar from any desired view \(v\) as follows:
\begin{smequation}
    \mathbf{I} = \text{AVA}(\hat{\IMug}, \IMug_{exp} -  \IMug, v, \mathbf{F}),
\end{smequation}
where \(\mathbf{I}\) represents the render of the avatar, and \(\mathbf{F}\) is the geometry generated during 3D face tracking for a monocular phone capture. $\text{AVA}$ corresponds to the inference process of the universal prior model. For more details about the universal prior model, please refer to \cite{cao2022authentic}.

%% file: 04_results.tex
\section{Results}
\label{sec:results}
\input{tables/supmetrics}
In this section, we introduce the dataset used in this paper, along with the baselines of our method. Subsequently, we present both quantitative and qualitative results of our method, comparing it to prior work. All phone-captured texture maps are generated using the mesh fitting procedure outlined in Authentic Volumetric Avatars (AVA) \cite{cao2022authentic}. Furthermore, we utilize AVA to render avatars based on the texture maps generated by all the methods, facilitating the calculation of image space metrics, including face identity similarity.
\vspace{-0.3cm}
\input{tables/faceid}

\paragraph{Training and Evaluation Data} We utilize a dataset comprising 12,543 neutral phone-captured texture maps to train \(\Gitw\), and 383 studio-captured texture maps for fine-tuning 
\(\Gitw\) to obtain \(\GMug\). Among the 383 maps, 83 are paired neutral texture maps 
\(\{\Iitw, \IGTM\}\) used for calculating the perceptual reconstruction loss, as described in \eq{percept_recons}. This paired dataset also serves for training the detail-preserving diffusion model. For quantitative and qualitative evaluation, we employ 10 paired phone-cum-studio captured texture maps and 31 phone-captured texture maps.

\vspace{-0.3cm}
\paragraph{Baselines} We compare our method to the following prior works on unpaired image-to-image translation. 1) AgileGAN \cite{song2021agilegan}, which utilizes an aligned StyleGAN latent space to stylize images, even with very few examples; 
2) Contrastive Unpaired Translation (CUT) \cite{cut}, which employs a patch-based contrastive loss in a learned feature space, enabling domain adaptation, even for single images; 
3) CycleGAN \cite{zhu2017cycleGAN}, which trains two generators based on a cycle-consistency loss to translate images between two domains.
Since AgileGAN \cite{song2021agilegan} has no code available, we implement it ourselves. We use publically available code for CUT \cite{cut} and \cite{zhu2017cycleGAN} for all experiments. 

\begin{figure}[ht]
    \centering
    \includegraphics[width=0.95\linewidth]{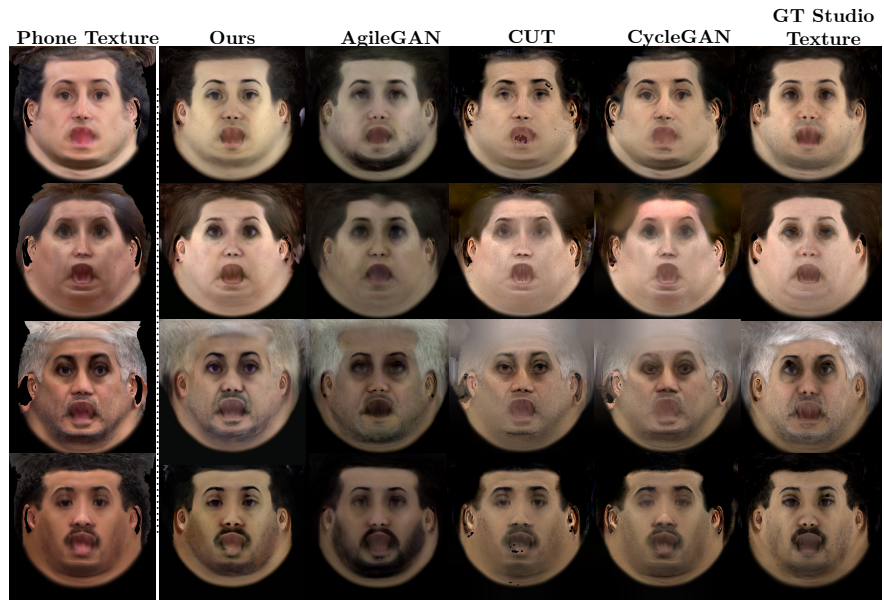}
    \caption{\small{\textbf{Comparisons with baselines on paired phone-cum-studio texture data} reveal that our method produces results closest to the ground truth. It achieves uniform studio-like lighting, well-reconstructed facial details visible in the phone capture, and effective inpainting of missing regions. In contrast, AgileGAN \cite{song2021agilegan} fails to preserve identity, while CUT \cite{cut} and CycleGAN \cite{zhu2017cycleGAN} introduce significant artifacts.    
    }}
    \vspace*{-5mm}
    \label{fig:baseline_comp_sup}
\end{figure}

\paragraph{Quantitative Results} In \tab{supervised_metrics}, we present the results of a quantitative evaluation comparing our method to the baselines using 10 paired texture maps captured with both phones and in a studio setting. The evaluation metrics include mean PSNR, SSIM, LPIPS \cite{perceptual}, and DISTS \cite{DISTS}. DISTS and LPIPS are perceptual metrics aligned with human judgment, while PSNR and SSIM are pixel-based metrics. As shown in \tab{supervised_metrics}, our method consistently outperforms the baseline across all metrics. The improvement is not limited to perceptual metrics but also extends to PSNR and SSIM, highlighting the efficacy of our proposed approach. It is worth noting that the relatively low PSNR and SSIM scores can be attributed to their sensitivity to small pixel shifts between the studio and phone-captured textures.

In \tab{faceid}, we display the face embedding distances, measured using \eq{faceid}, between avatars rendered from the phone-captured texture and the textures generated by various methods on the 31 unpaired phone captures. Our method exhibits the best preservation of facial identity.

\begin{figure}[t]
    \centering
    \includegraphics[width=0.95\linewidth]{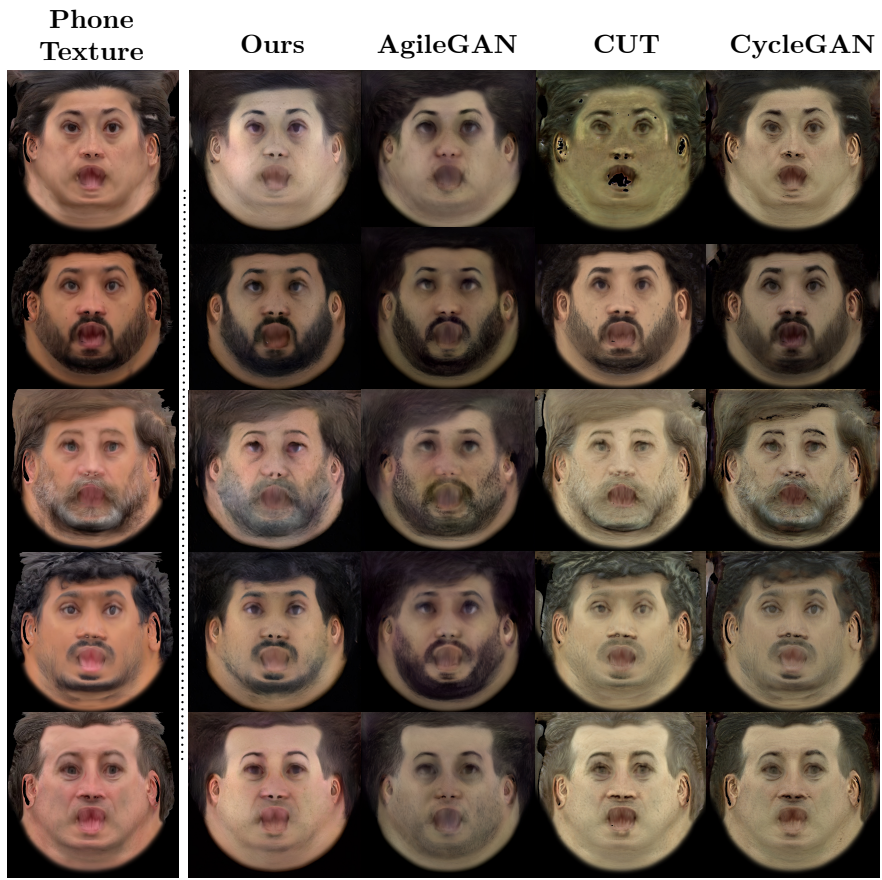}    
    \caption{\small{\textbf{Comparisons on Unpaired Phone-Captured Data}: In comparison to prior work, our method excels in generating texture maps with superior preservation of identity, enhanced photorealism in facial details, more uniform illumination, and improved inpainting of missing regions.
    }}
    \label{fig:baseline_comp_unsup}
    \vspace{-0.3cm}
\end{figure}

\begin{figure}
    \centering
    \includegraphics[width=0.80\textwidth]{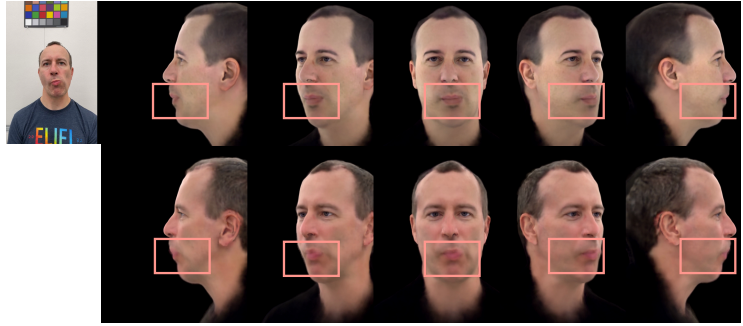}
    \caption{\small{\textbf{Avatar Reanimation.} The top row showcases a multiview render of an avatar generated using studio-like neutral and expression texture maps, created through our method as described in \sect{method_driving}. In contrast, the bottom row utilizes phone-captured texture maps. Notably, the lip region, highlighted with pink rectangles, appears more realistic and less blurry when using studio-like texture maps generated by our method.
    }}
    \label{fig:reanim}
    \vspace{-0.3cm}
\end{figure}

\begin{figure}[t]
    \centering
    \includegraphics[width=0.7\linewidth]{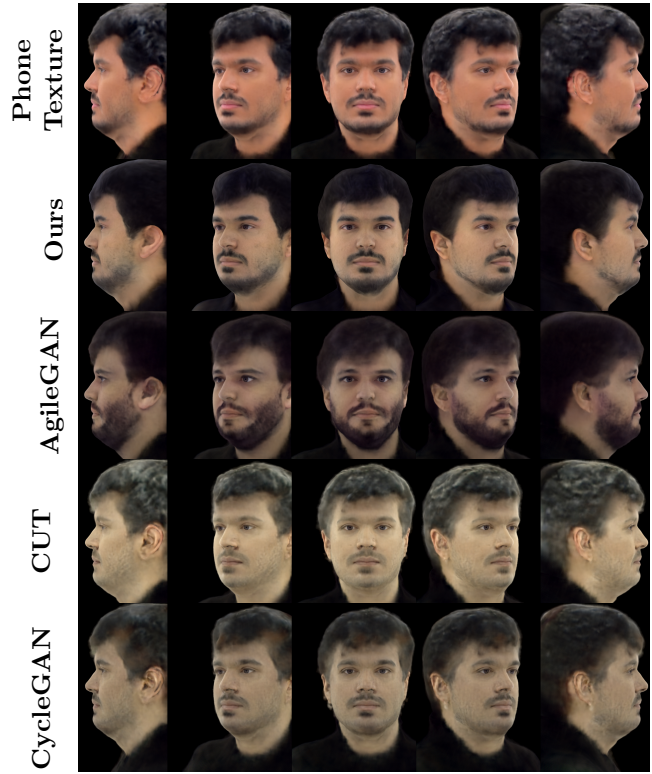} %
    
    \caption{\small{\textbf{Comparison of Avatars Generated Using \cite{cao2022authentic}}: In this comparison, we evaluate the quality of avatars generated by the Universal Prior Model (UPM) from AVA \cite{cao2022authentic}, using texture maps from various methods as input. Evidently, the avatars generated using texture maps from our method far exceed those generated by prior work.
    }}
    \label{fig:baseline_comp_unsup_rend}
    \vspace{-0.3cm}
\end{figure}

\paragraph{Qualitative Results} In \fig{baseline_comp_sup}, we provide some qualitative results on the paired test data. It is evident that our method generates a more plausible and photorealistic texture map compared to prior work, showcasing improvements in illumination transfer, facial details, and inpainting of missing regions.

In \fig{baseline_comp_unsup}, we present qualitative results comparing our method with prior work. While AgileGAN \cite{song2021agilegan} successfully changes the illumination to be studio-like and inpaints missing regions, it introduces a significant identity shift (quantitatively measured in \tab{faceid}) and lacks facial details. We attribute this to the use of the \(Z^+\) space \cite{song2021agilegan}, which may not be flexible enough for high-fidelity inversion, and the absence of identity-preserving constraints, both in architecture and optimization, during training. Due to its contrastive training paradigm, CUT \cite{cut} preserves identity better than AgileGAN but introduces significant artifacts. It is also unable to inpaint the missing regions around the ears and corners of the head. Similarly, like CUT \cite{cut}, CycleGAN \cite{zhu2017cycleGAN} also preserves identity better than AgileGAN \cite{song2021agilegan} but struggles with inpainting missing regions. The textures also contain numerous uncanny artifacts that are uncharacteristic of human skin. In contrast, our method generates a high-quality studio-illuminated texture map with accurate facial details and inpainted missing regions.

\input{tables/diffaba}
\paragraph{Reanimation Results} As explained in \sect{method_driving}, we can transform the phone expression texture maps \(\Iitw_{exp}\) to studio-like illumination using \eq{color_transform} and \eq{color_transform_tex}. We observe that utilizing studio-like illuminated expression texture maps \(\IMug_{exp}\) results in a modest improvement in the quality of reanimated avatars using the universal prior model from AVA \cite{cao2022authentic}. In \fig{reanim}, we present an example where the top row shows the render of an avatar using studio-like neutral and expression textures, while the bottom row displays a render using phone-captured neutral and expression textures. It is evident that the lip region appears more realistic and less blurry when reanimated with the studio-like neutral and expression textures compared to the phone-captured textures. We recommend that readers refer to our supplementary video for a more comprehensive comparison.

\subsection{Ablations}
\label{sec:aba}
In this section, we conduct ablation studies to evaluate the various components of our method.

\input{tables/wplusaba}
\paragraph{Details Conditioned Diffusion} 

We ablate the necessity of using facial details extracted from the phone capture texture map, in the form of an image gradient, to synthesize accurate facial details using a diffusion model. We explore three scenarios:

1) A Diffusion model that does not use any conditioning on phone-captured texture gradient (Vanilla Diffusion);
2) A Diffusion model that uses the phone-captured texture gradient only during inference but \textbf{not} during training;
3) Our model, which incorporates the phone-captured texture gradient during both training and inference.

\begin{figure}[t]
    \centering
    \includegraphics[width=0.85\textwidth]{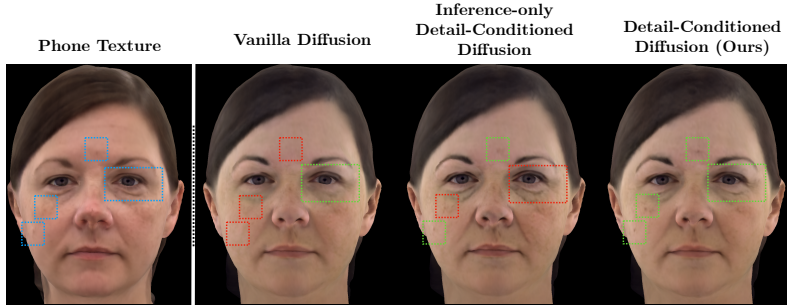}
    \caption{\small{\textbf{Ablation on Detail-conditioned Diffusion.} When no detail conditioning is applied, facial details visible in the phone capture are not reproduced. When detail conditioning is employed only during inference, undesirable shading effects are transferred to the generated texture map. Optimal results, with the most accurate and plausible reproduction of facial details, are achieved when detail conditioning is applied during both training and inference. Please refer to the text for further details.
    }}
    \label{fig:diffusion_ablation}
    \vspace{-0.2cm}
\end{figure}

In \fig{diffusion_ablation}, we present the results of each of the three scenarios. While the Vanilla Diffusion generates a realistic-looking facial texture, it fails to preserve facial details present in the phone-captured texture map. Prominent moles, marked by the blue boxes in the phone-captured texture map, are not generated by the Vanilla Diffusion model. When conditioning the reverse process using the phone-texture gradient only during inference, we observe that the model misses some details and transfers shading from the phone-captured texture map to the studio-lit texture map (marked by red boxes in \fig{diffusion_ablation}), which is undesirable. Ideally, we want facial details to be preserved while eliminating illumination-dependent effects, such as strong shading and shadows, from the studio-lit texture map. Our method conditions the diffusion model on the phone-texture gradient during both inference and training. As shown in \fig{diffusion_ablation}, this allows the model to learn to retain facial details while ignoring illumination-induced effects, such as strong shading, shadows, and specularities when synthesizing facial details. 
In \tab{diffaba_metrics}, we quantitatively validate each of the aforementioned diffusion model designs on paired data using LPIPS \cite{perceptual} and DISTS \cite{DISTS}. As can be seen, conditioning the diffusion model on phone-texture gradients both during training and inference gives the best results. 

\begin{figure}[t]
    \centering
    \includegraphics[width=0.80\linewidth]{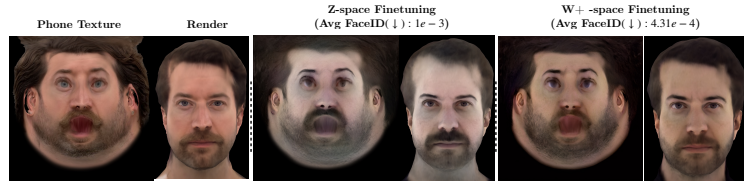}
    
    \caption{\small{\textbf{Qualitative \(W^+\) space ablation.} We observe that when \(\GMug\) is finetuned using the \(Z\) space, it fails to generalize to novel subjects and exhibits significant artifacts. In contrast, training \(\GMug\) in the \(W^{+}\) lends it far better generalization.    %
    }}
    \label{fig:wplus_ablation}
\end{figure}

\paragraph{\(W^+\) space vs. \(Z\) space for finetuning} We now evaluate the effectiveness of the \(W^{+}\) space over the standard \(Z\) space for fine-tuning \(\GMug\). For a quantitative comparison, we calculate the FaceID loss over the 31 unpaired evaluation texture maps and compute PSNR, SSIM, LPIPS \cite{perceptual}, and DISTS \cite{DISTS} over the 10 paired studio-cum-phone texture maps. Qualitative results, along with the average FaceID distance over unpaired data, are shown in \fig{wplus_ablation}, and the metrics on paired data are presented in \tab{wplus_metrics}. As evident, sampling in the \(W^{+}\) space generalizes much better to unseen identities, exhibiting significantly fewer artifacts and yielding superior results in terms of FaceID, PSNR, SSIM, LPIPS, and DISTS. We posit that this is because samples in the \(W^+\) space (i.e., samples from \(S_{W^{+}}\) space) are near-perfect inversions of the training data, making them more diverse than those generated from the Gaussian-distributed \(Z\) space. Consequently, as seen in \fig{wplus_ablation}, this leads to better generalization to unseen subjects. The model trained using the \(Z\) space exhibits uncanny artifacts in its results.

%% file: tables/supmetrics.tex
\begin{wraptable}{R}{5.0cm}
\vspace{-1.3cm}
\caption{\small{Quantitative results on paired phone-cum-studio captured texture maps. Our method outperforms prior work across all metrics. \mathcolorbox{pink}{\text{Best}} and \mathcolorbox{Yellow}{\text{Second Best}} scores are highlighted.}}
\scalebox{0.65}{
\begin{tabular}{lccccc}
\toprule
  \midrule
  Models   & PSNR $\uparrow$ & SSIM $\uparrow$ &LPIPS $\downarrow$ & DISTS $\downarrow$ \\
\midrule
Ours &
 \mathcolorbox{pink}{22.76} & \mathcolorbox{pink}{0.726} & \mathcolorbox{pink}{0.364} & \mathcolorbox{pink}{0.163}
 \\
 AgileGAN \cite{song2021agilegan} &
 18.05 &  \mathcolorbox{yellow}{0.657}  & 0.406  & 0.180 
 \\
 CUT \cite{cut} &
 \mathcolorbox{yellow}{21.440} &  0.642  & 0.402  & \mathcolorbox{yellow}{0.169} 
  \\
 CycleGAN \cite{zhu2017cycleGAN} &
  21.087 &  0.643  & \mathcolorbox{yellow}{0.400}  & 0.175 
 \\ 
\bottomrule
\end{tabular}}

\label{tab:supervised_metrics}
\vspace{-0.3cm}
\end{wraptable}

%% file: tables/faceid.tex
\begin{wraptable}{L}{5.0cm}
\small
\vspace{-0.5cm}
\caption{\small{FaceID results on unpaired phone captured texture maps. Our method better preserves identity than prior work without sacrificing the quality of the generated texture maps. \mathcolorbox{pink}{\text{Best}} and \mathcolorbox{Yellow}{\text{Second Best}} scores are highlighted. }
}
\scalebox{0.68}{
\begin{tabular}{lccccc}
\toprule
  \midrule
  Models  & Ours  & AgileGAN  & CUT  & CycleGAN  \\
\midrule
 FaceID & \mathcolorbox{pink}{4.31e-4} & \(1.36e-3\) & \(6.89e-4\) & \mathcolorbox{yellow}{5.19e-4} \\
\bottomrule
\end{tabular}}
\label{tab:faceid}

\end{wraptable}

%% file: tables/diffaba.tex
\begin{wraptable}{R}{5.0cm}
\begin{center}
\vspace{-0.8cm}
\caption{\small{\textbf{Detail conditioning ablation.}} We calculate quantitative metrics on the 10 paired phone-cum-studio captured texture maps. It is evident that utilizing detail conditioning during both training and inference yields the best performance.\mathcolorbox{pink}{\text{Best}} and \mathcolorbox{Yellow}{\text{Second Best}} scores are highlighted. 
}
\small
\scalebox{0.60}{
\begin{tabular}{lccc}
\toprule
  \midrule
  Models    &LPIPS $\downarrow$ & DISTS $\downarrow$ \\
\midrule
Vanilla Diffusion &
    0.383  & \mathcolorbox{yellow}{0.179} 
\\
Inference-only Details conditioning &
    \mathcolorbox{yellow}{0.376}  & 0.183 
\\
Training + Inference Details conditioning (Ours) &  \mathcolorbox{pink}{0.364} & \mathcolorbox{pink}{0.163}
 \\ 
\bottomrule
\end{tabular}}

\label{tab:diffaba_metrics}
\vspace{-0.4cm}
\end{center}
\end{wraptable}

%% file: tables/wplusaba.tex
\begin{wraptable}{R}{5.0cm}
\begin{center}
\caption{\small{\(W^{+}\) vs. \(Z\) Space ablation. } We utilize the 10 paired phone-cum-studio captured texture maps to compare the outcomes of training in the \(W^{+}\) and \(Z\) spaces. The results clearly demonstrate that training in the \(W^{+}\) space yields significantly better outcomes.
}
\small
\scalebox{0.65}{
\begin{tabular}{lccccc}
\toprule
  \midrule
  Models   & PSNR $\uparrow$ & SSIM $\uparrow$ &LPIPS $\downarrow$ & DISTS $\downarrow$ \\
\midrule
\(Z\) Space  &
 21.368 &  0.718  & 0.401  & 0.172
\\
\(W^{+}\) Space (Ours) &
  \mathcolorbox{pink}{22.76} & \mathcolorbox{pink}{0.726} & \mathcolorbox{pink}{0.364} & \mathcolorbox{pink}{0.163}\\
\bottomrule
\end{tabular}}

\label{tab:wplus_metrics}
\vspace{-0.5cm}
\end{center}
\end{wraptable}

%% file: 10_conclusion.tex
\vspace{-0.5cm}
\section{Conclusion and Limitations}
\label{sec:conclusion}
In this paper, we present a method for generating studio-like, high-quality avatars from monocular phone captures, using a StyleGAN2-based image-to-image translation and diffusion-based image upsampling. Experiments show the effectiveness of our approach in manipulating lighting, inpainting missing parts and generating facial details thus enabling the creation of complete, studio-lit textures for rendering high-quality avatars.
However, our method does have limitations. It struggles with input textures exhibiting extreme non-uniform lighting due to the constrained lighting conditions in our training data (we include examples in the supplementary). We also do not fine-tune the universal prior model to personalize the avatar based on phone capture data, as suggested in \cite{cao2022authentic}, thus the avatars lack personalized details for different facial expressions. Finally, our avatars are incomplete, featuring only the head. Future work involves extending the model to include the neck, shoulders, hands, and the entire body.